\documentclass[sigconf]{acmart}
\usepackage{bmpsize}
\usepackage{graphicx}
\usepackage{enumitem}
\usepackage{algorithm}
\usepackage{algpseudocode, algorithmicx}
\usepackage{bbm}
\usepackage{bm}
\usepackage{subfigure}
\usepackage{amsmath, amsthm}
\usepackage{commath}
\usepackage{stmaryrd}
\usepackage{float}
\usepackage[frozencache=true,cachedir=minted-cache]{minted} 
\usepackage{todonotes}

\AtBeginDocument{%
  \providecommand\BibTeX{{%
    \normalfont B\kern-0.5em{\scshape i\kern-0.25em b}\kern-0.8em\TeX}}}

\copyrightyear{2024}
\acmYear{2024}
\setcopyright{licensedusgovmixed}\acmConference[CIKM '24]{Proceedings of the 33rd ACM International Conference on Information and Knowledge Management}{October 21--25, 2024}{Boise, ID, USA}
\acmBooktitle{Proceedings of the 33rd ACM International Conference on Information and Knowledge Management (CIKM '24), October 21--25, 2024, Boise, ID, USA}
\acmDOI{10.1145/3627673.3679520}
\acmISBN{979-8-4007-0436-9/24/10}

\begin{document}

\title{Distilling Multi-Scale Knowledge for Event Temporal Relation Extraction}

\author{Hao-Ren Yao}
\authornote{Equal contributions.}
\email{hao-ren.yao@nih.gov}
\affiliation{%
  \institution{National Institutes of Health}
  \city{Bethesda}
  \state{MD}
  \country{USA}
}
\author{Luke Breitfeller}
\authornotemark[1]
\email{mbreitfe@andrew.cmu.edu}
\affiliation{%
  \institution{Carnegie Mellon University}
  \city{Pittsburgh}
  \state{PA}
  \country{USA}
}
\author{Aakanksha Naik}
\email{aakankshan@allenai.org}
\affiliation{%
  \institution{Allen Institute for AI}
  \city{Seattle}
  \state{WA}
  \country{USA}
}
\author{Chunxiao Zhou}
\email{chunxiao.zhou@nih.gov}
\affiliation{%
  \institution{National Institutes of Health}
  \city{Bethesda}
  \state{MD}
  \country{USA}
}
\author{Carolyn Rose}
\email{cprose@andrew.cmu.edu}
\affiliation{%
  \institution{Carnegie Mellon University}
  \city{Pittsburgh}
  \state{PA}
  \country{USA}
}

\renewcommand{\shortauthors}{H.-R. Yao et al.}

\begin{abstract}
Event Temporal Relation Extraction (ETRE) is paramount but challenging. Within a discourse, event pairs are situated at different distances or the so-called proximity bands. The temporal ordering communicated about event pairs where at more remote (i.e., ``long'') or less remote (i.e., ``short'') proximity bands are encoded differently. SOTA models have tended to perform well on events situated at either short or long proximity bands, but not both. Nonetheless, real-world, natural texts contain all types of temporal event-pairs. In this paper, we present \textbf{MulCo}: Distilling \textbf{Mul}ti-Scale Knowledge via \textbf{Co}ntrastive Learning, a knowledge co-distillation approach that shares knowledge across multiple event pair proximity bands to improve performance on all types of temporal datasets. Our experimental results show that MulCo successfully integrates linguistic cues pertaining to temporal reasoning across both short and long proximity bands and achieves new state-of-the-art results on several ETRE benchmark datasets.
\end{abstract}



\maketitle

\section{Introduction}
Event Temporal Relation Extraction (ETRE), which is a subtask of temporal reasoning, is a machine learning task where what is predicted is the order in which two event mentions from a text (also called an event pair) were depicted as happening in narrative time regardless of order of mention within the text. Though this ordering is a key component of understanding a sequence of events mentioned in text, a challenge arises when models must capture temporal cues for event pairs situated at disparate proximity bands with respect to one another.  An early ETRE benchmark dataset is the TimeBankDense (TB-Dense) corpus~\cite{cassidy2014annotation}, which features annotations for event pairs in articles from the TimeBankML news corpus~\cite{pustejovsky2003timeml}. This dataset biases heavily towards event pairs situated at ``short'' proximity bands,  i.e., situated within two sentences of one another within the text. Later work sought to complement TB-Dense with datasets that include annotations for event pairs situated at longer proximity bands (i.e., larger than three sentences): TDDiscourse-Auto (TDDAuto) and TDDiscourse-Manual (TDDMan). TDDAuto, curated for the most variety of short- and long-distance pairs among its set, designates itself as the complex real-world benchmark~\cite{naik2019tddiscourse}. The benefits of these corpora for ETRE training are that, in a real-world environment, queries on temporal relations may not be limited to event pairs situated within a specific proximity band. 

State-of-the-art ETRE models specializing in cues pertaining to one proximity band often fail to capture cues from other proximity bands. In particular, BERT-based context embeddings are well suited to Natural Language Processing (NLP) tasks that rely on short-distance linguistic cues~\cite{ning2019improved, han2019deep, wang2020joint, ballesteros2020severing}, but struggle to detect longer-range document properties like document structure~\cite{liu2021discourse, mathur2021timers, man2022selecting}. ETRE models that leverage Graph Neural Networks (GNN) to capture more expansive, long-distance structural cues better capture the way time is encoded across a whole document, important for corpora that include long proximity band event pairs~\cite{liu2021discourse}. Differences in performance between datasets with either primarily long, short, or an even mixture of proximity bands demonstrate a significant challenge in ETRE prediction: different information must be learned to account for event-pair ordering across proximity bands. TIMERS~\cite{mathur2021timers} is the first to explore the efficacy achieved through BERT and GNN fusion and achieves strong performance across existing temporal datasets. Still, its performance improvement substantially relies on a BERT-based context encoder, which suggests further improvement will require one of the following: \textbf{(1)} GNN models must be integrated to better tune for overcoming this challenge, or \textbf{(2)} a BERT model must be provided salient long-distance context information by another means.

The best-performing SOTA \cite{man2022selecting} follows option \textbf{(2)}, using strategic selection of additional long-distance sentence pairs across the document to augment BERT-based context encoding. This improvement is restricted by BERT's maximum input length, and therefore only a portion of the context for the longest-distance pairs can be used. Experimentation in Section~\ref{eval} demonstrates sub-optimal performance with this approach on datasets with only long-distance event pairs, e.g., TDDMan. One explanation is that there are distinct types of temporal cues required for full ETRE prediction, encompassing what is needed for both short and long proximity bands. A preliminary experiment in Table~\ref{bert-gnn-comparison} shows complementary subsets of event pairs correctly predicted by BERT and by GNN. Though BERT generally performs better than the GNN model, it fails to capture many long-distance pairs predicted correctly by GNN. GNN also demonstrates an advantage for processing short-distance event-pairs whose mentions are joined as dependent clauses, where the clause construction introduces variation in narrative time relative to mention ordering. Reporting verbs, for example, may introduce embedded clauses where an event has already happened, is predicted to happen, or is happening during the time of the report. This suggests the graph-based approach is able to more flexibly capture temporal relationships between clauses when mention order and narrative order differ. Our paper addresses the problem of capturing distinct cues associated with different proximity bands through exploitation of a \textbf{third} option: a single model formulation that combines knowledge across both long- and short-distance pairs, achieving better performance regardless of the composition of the corpus. We argue that combining event-ordering “knowledge'' between BERT and GNN to build an integrated model can achieve this high performance on both long- and short-distance event pairs. 

\begin{table}[]
\caption{Numbers of predictions that are only correctly predicted either by BERT or GNN. The detail of BERT and GNN models is in Section~\ref{implementation}, and dataset distribution is in Section~\ref{dataset}.}
\label{bert-gnn-comparison}
\centering
\begin{tabular}{l|cc}
\hline
\textbf{Test Set} & \textbf{TDDMan} & \textbf{TDDAuto} \\ \hline
BERT & 91 & 1065 \\
GNN & 170 & 105 \\ \hline
\end{tabular}
\end{table}

Therefore, we present the \textbf{MulCo}: \textbf{Mul}ti-Scale Contrastive Knowledge \textbf{Co}-Distillation model. MulCo performs knowledge distillation and transfer through a multi-scale \textbf{S}ubgraph-\textbf{A}ware \textbf{T}ransformer (SAT), using a contrastive objective to build a joint embedding space between BERT and GNN. As learning progresses, BERT and GNN embeddings function more similarly within the joint embedding space by maximizing the mutual information of both representations. To this end, we are no longer “ensembling'' but “distilling'' a diverse set of cues into a uniform “fused'' representation no longer specific to a type of event-pair or composition of trained corpus. Unlike prior approaches, which train a teacher first and then distill to the student, we formulate knowledge distillation as a single end-to-end contrastive learning paradigm. Knowledge, representation, and distillation between BERT- and GNN-based representations are simultaneously optimized, leading to a simpler, more effective, and higher-performing model. Comprehensive experiments on widely-used temporal relation extraction benchmarks validate each component of our model, which achieves new SOTA performance on event temporal relation extraction. Ablation studies further demonstrate the robustness and soundness of our proposed method.

\section{Related Work}
\subsection{Event temporal relation extraction} 
Many transformer-based temporal ordering models have been proposed in recent years ~\cite{ning2019improved, han2019deep, wang2020joint, ballesteros-etal-2020-severing, zhou2022rsgt}, and demonstrate strong results on short-distance event-pair corpora but not long-distance (see datasets in Section~\ref{dataset}). New work has attempted to bridge this gap using document-level Graph Neural Networks (GNNs) to capture syntactic, temporal, and rhetorical dependencies, which help to communicate time for long-distance pairs ~\cite{liu2021discourse,mathur2021timers,zhou2022rsgt}. Another approach uses reinforcement learning to retrieve optimal sentences for BERT-based training, and its implementation achieves the current SOTA in the field~\cite{man2022selecting}. This approach is more effective for short- than long-distance event pairs, as only a portion of the context for the longest-distance pairs can be meaningfully preserved in BERT embeddings. \\

\subsection{Knowledge Distillation} 
Our knowledge distillation builds on a existing method for compressing large pre-trained language models first introduced by ~\cite{hinton2015distilling}, using a simple student model to replicate the reasoning of a more sophisticated teacher model by minimizing KL-Divergence between output-class probability distributions of teacher and student. Other examples of traditional teacher-student knowledge distillation language models include ~\cite{sanh2019distilbert, sun2019patient, liang2020mixkd, liu2022multi}. ~\cite{zhang2018deep} introduced a variant where distillation was a multi-stage optimization process passing information mutually between teacher and student. Recently, contrastive learning has been combined with knowledge distillation ~\cite{tian2019crd} to compress large-scale pre-trained language models in a teacher-student representation space~\cite{sun2020contrastive, fu2021lrc}.

\begin{figure*}[htbp]
\centering
\centerline{\includegraphics[scale=0.22]{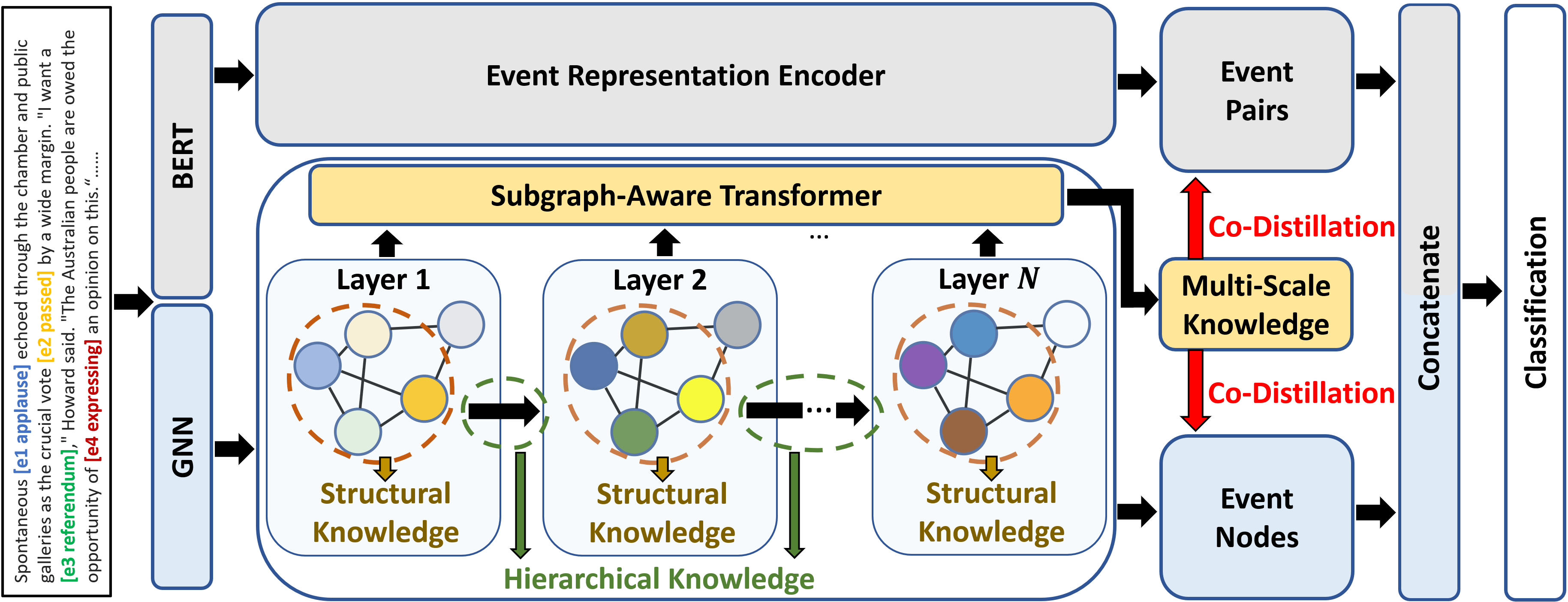}}
\caption{Overview of our proposed MulCo framework. First, we create event pairs and multi-scale event node representations from the input document using a BERT-based encoder and a GNN with a subgraph-aware transformer. Next, we apply contrastive knowledge distillation to align the event pairs with the multi-scale event node representations. Finally, we concatenate both representations and feed them into an event temporal relation classifier.}
\label{framework}
\end{figure*}

\section{Methodology}
In this section, we first review the elements of event temporal relation extraction relevant to our task and alternative event representations associated with BERT- and GNN-based approaches. Next, we describe our framework, starting from a vanilla knowledge distillation approach and expanding to structural and hierarchical distillation over event representations. We further propose a Subgraph-Aware Transformer (SAT) to reduce computation overhead and improve knowledge transfer between GNN and transformer-based language models. Last, we present MulCo (see Figure~\ref{framework}), an end-to-end training framework that jointly optimizes contrastive representation, mutual knowledge distillation, and classification objectives.

\subsection{Event Temporal Relation Extraction}
\subsubsection{Problem Definition}
Let document $\mathcal{D}$ consist of $M$ sentences $D=[S_1,\cdots, S_M]$, where each sentence contains a sequence of words, and $R$ is a set of temporal relations\footnote{Relation set depends on datasets listed in Section~\ref{dataset}.}, e.g., \textit{after}, \textit{before}, \textit{equal}, and \textit{vague}. Document $D$ has several event-pairs, each event-pair $(e^1_i, e^2_i)$ having a relation label $r_{i} \in R$. The event temporal relation extraction task is formulated as a multi-class classification problem to predict $r_{i}$ for a given event-pair $(e^1_i, e^2_i) \in D$. 

\subsubsection{BERT Event Representation Learning}
We employ a BERT-based pre-trained language model (BERT) $f_{\text{bert}}$ to encode event and event pair representations. Let $(e^1_i, e^2_i)$ be the $i$-th event pair, whose events $e^1_i, e^2_i$ are contained in sentences $S_p, S_q$ respectively. Without loss of generality, we follow prior work to extract a set of $2m-1$ sentences\footnote{$m$ will meet maximum BERT input length limitation.} to represent the event context:
\begin{gather}
\label{bert-host-neighbor}
D^1_i = \{S_{p-m+1},\cdots, S_{p+m-1}\} \\
D^2_i = \{S_{q-m+1},\cdots, S_{q+m-1}\}
\end{gather}
\noindent
where $D^1_i, D^2_i$ represents the event context of $e^1_i, e^2_i$ respectively. We then run BERT over $D^1_i$ and $D^2_i$ to obtain the BERT event representations $\bm{d^1_i}$ and $\bm{d^2_i}$:
\begin{gather}
    \bm{d^1_i} = f_{\text{bert}}(D^1_i) \\
    \bm{d^2_i} = f_{\text{bert}}(D^2_i)
\end{gather}
\noindent
where $\bm{d^1_i}, \bm{d^2_i} \in \mathbb{R}^{d_p}$, $d_p$ is the hidden dimension of $f_{\text{bert}}$. We then concatenate $\bm{d^1_i},\bm{d^2_i}$ into a BERT event pair representation $\bm{h^{\text{bert}}_i}$. \\

\subsubsection{Graph Event Representation Learning}
We strictly follow ~\cite{mathur2021timers} to represent the document as a syntactic-aware graph and a time-aware graph. \textbf{Syntactic-aware graph} $\mathcal{G}_{sg}=(\mathcal{V}_{sg},{\mathcal{E}_{sg}})$ represents the document's syntactic structure and contains three types of nodes $\mathcal{V}_{sg}$: document, sentence, and word. We construct four types of edge connections: \textbf{(i)} document-sentence, \textbf{(ii)} sentence-sentence, \textbf{(iii)} sentence-word, and \textbf{(iv)} word-word. \textbf{Time-aware graph} $\mathcal{G}_{tg}=(\mathcal{V}_{tg},{\mathcal{E}_{tg}})$ encodes document-level time-related information. It includes three types of nodes $\mathcal{V}_{tg}$: Document Creation Time (DCT), Time Expression (timex), and Word (corresponding to event word, with embeddings instantiated from BERT encoding), and edges: \textbf{(i)} DCT-timex, \textbf{(ii)} timex-timex, and \textbf{(iii)} word-timex. We use exactly the same way in~\cite{mathur2021timers} to initialize all node embeddings via BERT. Refer to~\cite{mathur2021timers} for more details. 

Without loss of generality, let graph $G=(\mathcal{V},A)$ be either syntactic or temporal graphs with nodes $\mathcal{V}$ and edge adjacency matrix $A$. Note, if not explicitly specified, $u_i$ denotes the $i$-th node in $G$. Let $u^1_i, u^2_i \in \mathcal{V}$ denote nodes corresponding to events $e^1_i$, $e^2_i$. We obtain the $l$-th layer event node representation $\bm{h^{1(l)}_i}, \bm{h^{2(l)}_i}$ for $u^1_i, u^2_i$ via a $L$ layer Graph Neural Network (GNN) $f_\text{gnn}$:
\begin{align}
    \bm{h^{1(l)}_i} &= f_\text{gnn}(\bm{h^{1(l-1)}_i}, A) \\
    \bm{h^{2(l)}_i} &= f_\text{gnn}(\bm{h^{2(l-1)}_i}, A)
\end{align}
\noindent
where $\bm{h^{1(l)}_i},\bm{h^{2(l)}_i} \in \mathbb{R}^{d_g}$, and $d_g$ is the hidden dimension of $f_\text{gnn}$. For each event node, we fuse their $L$ layer node representations into the final event node representation as follows:
\begin{align}
    \bm{h^{1}_i} &= f_{\theta}([\bm{h^{1(1)}_i},\dots,\bm{h^{1(l)}_i}]) \\
    \bm{h^{2}_i} &= f_{\theta}([\bm{h^{2(1)}_i},\dots,\bm{h^{2(l)}_i}])
\end{align}
\noindent
where $f_{\theta}$ is a single layer multilayer perceptron (MLP). Finally, we concatenate $\bm{h^{1}_i}, \bm{h^{2}_i}$ into an event pair representation $\bm{h^{\text{gnn}}_i}$.

\subsection{Knowledge Distillation}
\subsubsection{Vanilla Distillation} 
Let $f_{\alpha}$ and $f_{\beta}$ be the fully-connected layers that output predicted logits for BERT and GNN respectively. Given $N$ event pairs $\{(e^1_i, e^2_i) \mid i=1,\dots,N\}$, we can transfer knowledge from GNN to BERT by a knowledge distillation objective from~\cite{hinton2015distilling} in order to minimize the KL-Divergence (KL) between class output probability distributions from the logits of BERT and GNN event pair representations:
\begin{equation}
\label{vkd}
    \mathcal{L}_\text{KD} = \sum^N_{i=1} \text{KL}(\sigma(f_{\alpha}(\bm{h^{\text{bert}}_i})), \sigma(f_{\beta}(\bm{h^{\text{gnn}}_i})))
\end{equation}
\noindent
where $\sigma(\cdot)$ denotes a softmax function. A limitation of such distillation is that it only transfers knowledge that contributes to successful predictions of event temporal relations~\cite{tian2019crd}. Generally, it overlooks structural knowledge from representation learning, which leads to performance bottlenecks. Hence, we propose to distill hierarchical structural knowledge in the embedding space of the GNN to BERT. 

\subsubsection{Structural Distillation} 
We are interested in transferring knowledge, especially rich structural information encoded within the graph, e.g., edge relation and topology, from the GNN into BERT. Theoretically, a node's k-hop neighborhood contains expressive multi-hop topology that is powerful for representation of structural information~\cite{feng2022powerful}. We denote $\mathcal{G}_{u_i}$ as the set of all nodes in the k-hop neighborhood of node $u_i$:
\begin{equation}
\label{subgraph-aug}
    \mathcal{G}_{u_i} = \{h_i\} \cup \{h_j \mid j \in \mathcal{N}^{k}_{i} \} 
\end{equation}
\noindent
where $\mathcal{N}^{k}_{i}$ denotes a k-hop neighborhood of node $u_i$. As~\cite{tian2019crd} suggest, contrastive representation learning improves structural knowledge distillation from a teacher network in that it captures structural information from the teacher's representation space:
\begin{equation}
\label{loss-cl}
    \ell_\text{cl}(t,s) =-log\frac{e^{\text{sim}(t,s)/\tau}}{\sum_q\mathbbm{1}_{[q \ne t]}e^{\text{sim}(t,q)/\tau}}
\end{equation}
\noindent
where $t,s$ are representations from the teacher and student, $\text{sim}(\cdot)$ denotes cosine similarity\footnote{In fact, it can be any similarity function, e.g., dot product.}, $\mathbbm{1}_{[q \ne t]} \in \{0,1\}$ is an indicator to evaluate if $q \ne t$ where $q$ denotes other representations from the training data, and $\tau$ is a temperature parameter. When $t,s$ are in different dimensions, we employ a simple linear transformation to project them into the same dimension. Given an event pair $(e^1_i, e^2_i)$, we can transfer structural knowledge from the GNN to BERT by using a structural distillation objective to minimize the contrastive loss in Eq.\ref{loss-cl} between an event pair's BERT-based representation and all node representations within the k-hop neighborhood of event node $u^1_i$ and $u^2_i$
\begin{equation}
\label{skd}
    \mathcal{L}_\text{SD} = \sum^N_{i=1}\sum^{\abs{\mathcal{G}_i}}_{j=1}\ell_{cl}(\bm{h^\text{bert}_i},v_j)
\end{equation}
\noindent
where $\mathcal{G}_i$ denotes the set of all node representations in the k-hop neighborhood of event node $u^1_i$ and $u^2_i$ such that $\mathcal{G}_i = \mathcal{G}_{u^1_i} \cup \mathcal{G}_{u^2_i}$. 

\subsubsection{Hierarchical Distillation} When stacking $L$ layer GNNs, we are able to distill hierarchical structural knowledge from GNN to BERT, which greatly improves long-distance event modeling. We extend the structural distillation objective Eq.\ref{skd} to a hierarchical distillation objective as follows: 
\begin{equation}
\label{hkd}
    \mathcal{L}_\text{HD} = \sum^N_{i=1}\sum^L_{l=1}\sum^{\abs{\mathcal{G}^l_i}}_{j=1}\ell_\text{cl}(\bm{h^\text{bert}_i},v_j)
\end{equation}
\noindent
where $\mathcal{G}^l_i$ denotes the set of all  $l$-th layer node representations in the k-hop neighborhood of event nodes $u^1_i$ and $u^2_i$ such that $\mathcal{G}^l_i = \mathcal{G}^l_{u^1_i} \cup \mathcal{G}^l_{u^2_i}$. 

Naturally, larger number of hops and layers increase the distillation capacity with more knowledge transfer from both structural and hierarchical context. However, it might degrade the performance by over-distillation that leads to over-fitting. We investigate the knowledge distillation behavior under various of setting of k-hop and number of layers in Section~\ref{k-hops-performance}.

\subsection{Multi-Scale Knowledge Co-Distillation}
\subsubsection{Subgraph-Aware Transformer} 
As described above, hierarchical distillation effectively transfers global knowledge from a GNN into BERT. However, such distillation leads to a large computational overhead. However, the complexity of the number of hierarchical distillation targets is $\mathcal{O}(N \times L \times K)$. The hierarchical distillation will become intractable when increasing hops of neighborhood $k$ and depth of GNN layer $L$. To address this issue, we propose a \textbf{Subgraph-Aware Transformer (SAT)} $f_\text{sat}$ and perform two-stage fusion on a multi-layer k-hop neighborhood. Given a set of $d_g$-dimension node representations $\mathcal{G}=\{h_1,\cdots,h_m\} \in \mathcal{R}^{m \times d_g}$, \textbf{SAT} computes pairwise self-attentive node representations $\mathcal{G}_a \in \mathcal{R}^{m \times d_g}$ via scaled dot-product attention followed by mean-pooling (Pool) to produce a single integrated node representation $\mathcal{H} \in \mathcal{R}^{d_g}$:
\begin{align}
         \mathcal{H} &= f_\text{sat}(\mathcal{G}) \\
         &= \text{Pool}(\sigma(\frac{\mathcal{G}W_Q({\mathcal{G}W_K)}^{\intercal}}{\sqrt{d_a}})\mathcal{G}W_V)
\end{align}
\noindent
where $W_Q, W_K$, and $W_V \in \mathcal{R}^{d_g \times d_a}$ are attention weights, and $\sigma$ denotes a softmax function. 

\subsubsection{Multi-Scale Distillation} 
Given $\mathcal{G}^l_i$, a set of $l$-th layer node representations in the k-hop neighborhood of event node $u^1_i$ and $u^2_i$, we obtain the fused structural representation $\bm{H^l_i}$ of $\mathcal{G}^l_i$ with \textbf{SAT}:
\begin{equation}
    \bm{H^l_i} = f_{\text{sat}}(\mathcal{G}^l_i)
\end{equation}
We aggregate all fused structural representations pulled from the $L$ layer of event node $u^1_i$ and $u^2_i$ into a set and employ \textbf{SAT} to obtain the final layer-wise fused hierarchical representation $\bm{\mathcal{H}_i}$:
\begin{equation}
    \bm{\mathcal{H}_i} = f_{\text{sat}}([\bm{H^1_i},\cdots,\bm{H^L_i}])
\end{equation}
$\mathcal{H}_i$ can be treated as a multi-scale context representation that characterizes event pair relations combining both structural and hierarchical aspects. Furthermore, its single and compact representation decreases the number of targets in the distilled model. Therefore, we can re-formulate the hierarchical distillation in Eq.\ref{hkd} as Multi-Scale Distillation: 
\begin{equation}
\label{mkd}
    \mathcal{L}_\text{MD} = \sum^N_{i=1}\ell_\text{cl}(\bm{h^\text{bert}_i},\bm{\mathcal{H}_i})
\end{equation}
\noindent
To this end, we can efficiently transfer multi-scale knowledge from the GNN to BERT. The number of distillation targets is now independent of the number of hops and layers. The complexity is reduced to $\mathcal{O}(N)$. 

\subsubsection{Contrastive Co-Distillation}
In general, knowledge distillation consists of two-phase training (1) train the teacher's network on a task objective; (2) train the student network on both task and knowledge distillation objectives. This one-way knowledge distillation cannot fully explore shared knowledge between the two networks to produce better generalization~\cite{zhang2018deep, li2020towards}. We propose a more powerful approach, \textbf{Contrastive Co-Distillation (CoD)}\footnote{When distilling from a syntactic-aware graph and a time-aware graph, we add their corresponding CoD loss.}, which combines contrastive representation learning and stop gradient operation. In particular, we concatenate BERT and GNN outputs $\bm{h^\text{bert}_i},\bm{h^\text{gnn}_i}$ into $\bm{h_i}$ for classification and re-formulate Eq.\ref{mkd} as Eq.~\ref{cod} to form our \textbf{MulCo} objective:
\begin{gather}
    \mathcal{L}_\text{CoD} = \sum^N_{i=1}\ell_\text{cl}(\bm{h^\text{bert}_i},\hat{\bm{\mathcal{H}_i}}) + \ell_\text{cl}(\bm{\mathcal{H}_i},\hat{\bm{h^\text{bert}_i}}) \label{cod} \\
    \mathcal{L}_\text{CLF} = \sum^N_{i=1}\ell_\text{ce}(f_{\phi}(\bm{h_i}), r_i) \\
    \mathcal{L}_\text{MulCo} = \mathcal{L}_\text{CoD} + \mathcal{L}_\text{CLF}
\end{gather}
\noindent
where $\hat{\cdot}$ stands for the stop gradient operator~\cite{chen2021exploring} that sets the input variable to a constant, $f_{\phi}$ is a fully-connected layer that outputs the class prediction, $r_i$ is the temporal relation for the $i$-th event pair, and $\ell_\text{ce}$ is the cross-entropy loss. Instead of a two-phase training or a multi-step alternative optimization as in~\cite{zhang2018deep}, we optimize the whole framework using a single loss $\mathcal{L}_\text{MulCo}$ with end-to-end training. The concatenated output $h_i$ acts as an ensemble to deliver comprehensive multi-modal information. As such, \textbf{MulCo} can simultaneously perform mutual distillation and model optimization on BERT and GNN toward classification objectives. To verify the advantages of end-to-end co-distillation, we investigate the performance on different KD frameworks in Section~\ref{eval} and provide the theoretical interpretation of end-to-end training in Section~\ref{stop-gradient}.

\section{Experiments}
\subsection{Datasets statistics and Splits}
\label{dataset}
To demonstrate our success at overcoming specificity either to short-distance or long-distance pairs, we select four widely-used event temporal relation extraction datasets in Table~\ref{stats} as benchmarks\footnote{Appendix~\ref{appendix-iaa} argues and discusses the potential question regarding performance versus the Inter-Annotator Agreement (IAA) for benchmark datasets.}: TimeBank-Dense (TB-Dense)~\cite{cassidy2014annotation}, MATRES~\cite{ning2018multi}, and TDDiscourse~\cite{naik2019tddiscourse}, which is further subdivided into TDDMan and TDDAuto. TB-Dense and MATRES largely contain short-distance event pairs, where temporal relations are either located in the same or adjacent sentence. TDDMan only contains long-distance event pairs, and TDDAuto contains a mixture of short- and long-distance event pairs. For our evaluations, we follow the same data splits (train/val/test) and evaluate under F1 score\footnote{The F1 score calculated on TB-Dense and MATRES might cause confusion compared to other literature. We detail this in the Appendix~\ref{appendix-f1}.} as in all prior works. Table~\ref{relations} lists all possible temporal relations in each dataset.

\begin{table}[htb]
\caption{Dataset statistics on Train/Val/Test splits and the number of relation types (labels).}
\label{stats}
\centering
\begin{tabular}{l|ccc|c}
\hline
\textbf{Dataset}        & \textbf{Train}  & \textbf{Val}   & \textbf{Test}  & \textbf{Labels} \\ \hline
TDDMan         & 4,000  & 650   & 1,500 & 5      \\
TDDAuto        & 32,609 & 4,435 & 4,258 & 5      \\ 
MATRES         & 231    & 25    & 20    & 4      \\ 
TB-Dense       & 4,032  & 629   & 1,427 & 6      \\ \hline
\end{tabular}
\end{table}

\begin{table}[htb]
\caption{Temporal relation types summary on each dataset.}
\label{relations}
\centering
\begin{tabular}{l|l}
\hline
\textbf{Dataset} & \multicolumn{1}{c}{\textbf{Temporal Relations}} \\ \hline
TDDAuto & \textit{After, Before, Simultaneous, Includes, Is included} \\
TDDMan & \textit{After, Before, Simultaneous, Includes, Is included} \\
MATRES & \textit{After, Before, Equal, Vague} \\
TB-Dense & \textit{After, Before, Simultaneous, Includes, Is included, Vague} \\ \hline
\end{tabular}
\end{table}

\subsection{Baselines}
To fairly compare MulCo against recent SOTAs, we select commonly-used baselines: \textbf{SP+ILP}~\cite{ning2017structured}, \textbf{BiLSTM}~\cite{cheng2017classifying}, \textbf{BiLSTM+MAP}~\cite{han2019joint}, and \textbf{DeepSSVM}~\cite{han2019deep}. We also select \textbf{Longformer}~\cite{beltagy2020longformer}, \textbf{Reformer}~\cite{kitaev2020reformer}, and \textbf{BigBird}~\cite{zaheer2020big} long-document pre-trained language models\footnote{We adopt the implementation in~\cite{man2022selecting}.} to evaluate MulCo long-document distillation. From prior GNN SOTAs, we include \textbf{UCGraph}~\cite{liu2021discourse}, \textbf{TIMERS}~\cite{mathur2021timers}, and \textbf{RSGT}~\cite{zhou2022rsgt}. Finally, we look at the current best-performing approache \textbf{SCS-EERE}~\cite{man2022selecting} and \textbf{Unified-Framework (UF)}~\cite{huang2023more}. To test our knowledge distillation, we compare MulCo with three common training schemes: vanilla knowledge distillation (\textbf{VKD})~\cite{hinton2015distilling}, mutual distillation, (\textbf{DML})~\cite{zhang2018deep}, and contrastive representation distillation (\textbf{CRD})~\cite{tian2019crd}. In our setting, \textbf{VKD} optimizes a one-way KL objective (Eq.\ref{vkd}) on the class probability distribution; \textbf{DML} optimizes two-way KL objectives on both teacher and student class probability distributions and a multi-scale distillation objective (MD in Eq.\ref{mkd}); and \textbf{CRD} only optimizes an MD objective. All previous distillations require two-phase training.

\subsection{Implementation Details}
\label{implementation}
We use vanilla \textbf{BERT}~\cite{Devlin2019bert} and \textbf{RoBERTa}~\cite{liu2019roberta} as our pre-trained language models with the two event context modeling strategies we have adopted: \textbf{host-sentence}, e.g., only including the sentence with the event, and \textbf{neighbor-sentence}, e.g., including a set of sentences immediately before and after the event sentence, for which decent baseline performance is demonstrated in~\cite{man2022selecting}. Specifically, we initialize BERT and RoBERTa from ``bert-based-uncased'' and ``roberta-base'' respectively with hidden size fixed to 768. Note that distilling knowledge into long-document pre-trained language models is not our motivation as we notice their sub-optimal performance on long-distance event pairs compared to vanilla BERT with optimal sentence selection baseline~\cite{man2022selecting}. All the model parameters for ``bert-based-uncased'' and ``roberta-base'' remain the default setting.

\begin{table}[htb]
\caption{Hyperparameter Settings.}
\label{hyperparameters}
\centering
\begin{tabular}{l|c}
\hline
\textbf{Parameters} & \textbf{Value Range} \\ \hline
BERT dimension & 768 \\
GNN dimension & \{64, 128, 256\} \\
RGAT attention dimension & \{64, 128, 256\} \\
SAT dimension & 768 \\
Embedding space for cl & 2048 \\
K-hops & \{1, 2, 3, 4, 5\} \\
Number of GNN layers & \{1, 2, 3, 4, 5\} \\
Dropout & \{0.1, 0.2, ..., 0.8, 0.9\} \\
Learning Rate & Fixed 1e-5 \\
Batch size & \{8, 16, 32\} \\
Temperature in CL & \{0.01, 0.02, ..., 0.8, 0.9 \} \\
Temperature in KD & Fixed 0.1 \\ \hline
\end{tabular}
\end{table}

As different GNNs exhibit distinct characteristics that provide different knowledge, we distill knowledge from the different choices of GNNs to BERT: \textbf{Graph Convolution Network (GCN)}~\cite{kipf2016semi}, \textbf{Relational Graph Convolution Network (RGCN)}~\cite{schlichtkrull2018modeling}, and \textbf{Relational Graph Attention Network (RGAT)}~\cite{busbridge2019relational}. We employ a two-layer multilayer perceptron (MLP) with ReLu activation as event temporal relation classifier. We tune hyper-parameters listed in Table~\ref{hyperparameters} on the validation set. We use the Adam~\cite{kingma2014adam} optimizer with a fixed learning rate of 0.0001. We use SpaCy 3.4 to produce document and sentence dependency parsing, PyTorch Geometric~\cite{Fey/Lenssen/2019} 2.0.4 to implement all GNN models and PyTorch version of Huggingface Transformers for BERT and RoBERTa. All experiments are written in Python and run with a random seed setup on DGX-A100 with 40GB GPU memory. Optimal parameters and random seed for each dataset are reported in Table~\ref{detail-params}. 


\begin{table}[]
\caption{Optimal hyperparameters that are selected for each dataset.}
\label{detail-params}
\centering
\begin{tabular}{l|cccc}
\hline
\textbf{Parameters} & \begin{tabular}[c]{@{}c@{}}\textbf{TDD}\\ \textbf{Man}\end{tabular} & \begin{tabular}[c]{@{}c@{}}\textbf{TDD}\\ \textbf{Auto}\end{tabular} & \textbf{MATRES} & \begin{tabular}[c]{@{}c@{}}\textbf{TB-}\\ \textbf{Dense}\end{tabular} \\ \hline
BERT dimension & 768 & 768 & 768 & 768 \\
GNN dimension & 256 & 256 & 256 & 128 \\
RGAT attention dimension & 256 & 256 & 256 & 128 \\
SAT dimension & 768 & 768 & 768 & 768 \\
Embedding space for cl & 2048 & 2048 & 2048 & 2048 \\
K-hops & 1 & 2 & 1 & 1 \\
Number of GNN layers & 2 & 3 & 1 & 1 \\
Dropout & 0.1 & 0.1 & 0.3 & 0.1 \\
Learning Rate & 1e-5 & 1e-5 & 1e-5 & 1e-5 \\
Batch size & 16 & 32 & 16 & 16 \\
Temperature in CL & 0.1 & 0.04 & 0.9 & 0.9 \\
Temperature in KD & 0.1 & 0.1 & 0.1 & 0.1 \\ 
Random Seed & 2513 & 2513 & 1103 & 1103 \\ \hline
\end{tabular}
\end{table}

\subsection{Evaluation Results}
\label{eval} 
\subsubsection{Vanilla Knowledge Distillation} 
We first investigate the extent to where GNNs provide “knowledge'' that improves BERT on long-distance event-ordering. Table~\ref{distill-eval} evaluates a baseline BERT model with various KD from a GNN. It is evident that distilling knowledge from a GNN removes the document-length barrier of BERT context encoding. As Table~\ref{distill-eval} shows, vanilla KD improves host-sentence BERT baselines, rendering them competitive with long-document language models (See Long-Document LM in Table~\ref{overall-eval}) on long-distance event pair datasets (TDDMan and TDDAuto). Hence, we can conclude that a GNN does provide ``knowledge'' for BERT that improves long-distance event pair modeling. 

\begin{table}[]
\caption{F1-score comparison for different KD methods. For each KD method, we report the best score from the best distilled GNN listed in Section~\ref{backbone-architecture}.}
\label{distill-eval}
\centering
\begin{tabular}{lccccc}
\hline
\multicolumn{1}{l|}{\textbf{BERT Variants}} & \textbf{No KD} & \textbf{VKD} & \textbf{DML} & \textbf{CRD} & \textbf{MulCo} \\ \hline
\multicolumn{6}{c}{\textbf{TDDMan}} \\ \hline
BERT-Host & 37.5 & 43.0 & 45.4 & 44.1 & 47.3 \\
BERT-Neighbor & 44.7 & 44.0 & 46.2 & 47.9 & 49.1 \\
RoBERTa-Host & 37.1 & 49.1 & 51.3 & 50.1 & 51.5 \\
RoBERTa-Neighbor & 44.5 & 47.9 & 51.2 & 50.2 & \textbf{55.1} \\ \hline
\multicolumn{6}{c}{\textbf{TDDAuto}} \\ \hline
BERT-Host & 62.3 & 67.4 & 69.3 & 68.8 & 70.3 \\
BERT-Neighbor & 62.4 & 66.9 & 65.7 & 66.3 & 68.3 \\
RoBERTa-Host & 61.6 & 73.3 & 72.7 & 76.9 & 76.0 \\
RoBERTa-Neighbor & 68.9 & 69.7 & 72.1 & 72.6 & \textbf{77.1} \\ \hline
\multicolumn{6}{c}{\textbf{MATRES}} \\ \hline
BERT-Host & 78.1 & 85.7 & 84.9 & 87.5 & 85.7 \\
BERT-Neighbor & 80.2 & 83.5 & 82.7 & 84.9 & 86.0 \\
RoBERTa-Host & 78.4 & 87.9 & 87.5 & 88.2 & 89.7 \\
RoBERTa-Neighbor & 79.4 & 85.3 & 86.8 & 87.5 & \textbf{90.4} \\ \hline
\multicolumn{6}{c}{\textbf{TB-Dense}} \\ \hline
BERT-Host & 62.2 & 78.7 & 80.2 & 78.1 & 83.5 \\
BERT-Neighbor & 77.5 & 77.4 & 79.0 & 77.3 & 81.0 \\
RoBERTa-Host & 61.9 & 79.5 & 79.8 & 78.3 & 84.2 \\
RoBERTa-Neighbor & 76.7 & 78.7 & 83.5 & 82.1 & \textbf{85.6} \\ \hline
\end{tabular}
\end{table}

\subsubsection{Multi-Scale Distillation} 
Next, we investigate the extent to which multi-scale knowledge provides compact and comprehensive representations that allow BERT to capture structural and hierarchical context cues. We compare Vanilla KD with Multi-Scale Distillation (\textbf{DML}, \textbf{CRD}, and \textbf{MulCo}). Table~\ref{distill-eval} shows all BERT baselines substantially improve on both long- and short-distance event temporal relation extraction when passed through multi-scale knowledge distillation. They outperform or perform competitively with current SOTAs across comparisons in Table~\ref{overall-eval}. This confirms the efficacy of multi-scale knowledge and demonstrates the Subgraph-Aware Transformer fuses structural and hierarchical knowledge into a compact representation. 

\subsubsection{End-to-End Co-Distillation} 
Finally, to see the validity of our end-to-end formulation we measure the performance of MulCo's objective against various KD frameworks. We see in Table~\ref{distill-eval} that end-to-end co-distillation \textbf{MulCo} achieves better and more efficient mutual distillation without alternative optimization (\textbf{DML}) and two-phase distillation on representation (\textbf{KD} and \textbf{CRD}). With this distillation, MulCo establishes new SOTA performance in event temporal relation extraction as in Table~\ref{overall-eval}, confirming our hypotheses on GNN knowledge, distillation to BERT, and efficient end-to-end co-distillation. We observe that neighbor-sentence event modeling also has benefits in knowledge distillation, suggesting that the inclusion of more sentences for event pair modeling benefits contextualized learning.

\begin{table}[]
\caption{F1-score comparison on all baselines and SOTAs, best scores bolded. For MulCo, we select the best-performing variants from Table~\ref{distill-eval}.}
\label{overall-eval}
\centering
\begin{tabular}{lcccc}
\hline
\multicolumn{1}{l|}{\textbf{Model}} & \multicolumn{1}{c|}{\textbf{TDDMan}} & \multicolumn{1}{c|}{\textbf{TDDAuto}} & \multicolumn{1}{c|}{\textbf{MATRES}} & \textbf{TB-Dense} \\ \hline
\multicolumn{5}{c}{\textbf{Previous Baselines}} \\ \hline
\multicolumn{1}{l|}{SP+ILP} & \multicolumn{1}{c|}{23.8} & \multicolumn{1}{c|}{46.1} & \multicolumn{1}{c|}{76.3} & 58.4 \\
\multicolumn{1}{l|}{BiLSTM} & \multicolumn{1}{c|}{24.3} & \multicolumn{1}{c|}{51.8} & \multicolumn{1}{c|}{59.5} & 48.4 \\
\multicolumn{1}{l|}{BiLSTM+MAP} & \multicolumn{1}{c|}{41.1} & \multicolumn{1}{c|}{57.1} & \multicolumn{1}{c|}{75.5} & 64.5 \\
\multicolumn{1}{l|}{DeepSSVM} & \multicolumn{1}{c|}{41.0} & \multicolumn{1}{c|}{58.8} & \multicolumn{1}{c|}{-} & 63.2 \\ \hline
\multicolumn{5}{c}{\textbf{Long-Document LM}} \\ \hline
\multicolumn{1}{l|}{Reformer} & \multicolumn{1}{c|}{43.7} & \multicolumn{1}{c|}{65.9} & \multicolumn{1}{c|}{-} & - \\
\multicolumn{1}{l|}{BigBird} & \multicolumn{1}{c|}{43.3} & \multicolumn{1}{c|}{65.3} & \multicolumn{1}{c|}{-} & - \\
\multicolumn{1}{l|}{Longformer} & \multicolumn{1}{c|}{44.2} & \multicolumn{1}{c|}{66.8} & \multicolumn{1}{c|}{-} & - \\ \hline
\multicolumn{5}{c}{\textbf{Recent SOTAs}} \\ \hline
\multicolumn{1}{l|}{UCGraph} & \multicolumn{1}{c|}{43.4} & \multicolumn{1}{c|}{61.2} & \multicolumn{1}{c|}{-} & 59.1 \\
\multicolumn{1}{l|}{TIMERS} & \multicolumn{1}{c|}{45.5} & \multicolumn{1}{c|}{71.1} & \multicolumn{1}{c|}{82.3} & 67.8 \\
\multicolumn{1}{l|}{RGST} & \multicolumn{1}{c|}{-} & \multicolumn{1}{c|}{-} & \multicolumn{1}{c|}{82.2} & 68.7 \\
\multicolumn{1}{l|}{UF} & \multicolumn{1}{c|}{-} & \multicolumn{1}{c|}{-} & \multicolumn{1}{c|}{82.6} & 68.1 \\
\multicolumn{1}{l|}{SCS-EERE} & \multicolumn{1}{c|}{51.1} & \multicolumn{1}{c|}{76.7} & \multicolumn{1}{c|}{83.4} & - \\ \hline
\multicolumn{1}{l|}{\textbf{MulCo}} & \multicolumn{1}{c|}{\textbf{55.1}} & \multicolumn{1}{c|}{\textbf{77.1}} & \multicolumn{1}{c|}{\textbf{90.4}} & \textbf{85.6} \\ \hline
\end{tabular}
\end{table}

\section{Discussion}
\label{discussion}
\subsection{Analysis on MulCo knowledge transfer}
Here, we more deeply analyze MulCo performance on TDDAuto\footnote{The evaluation works mainly on TDDAuto since it contains the most variety of short- and long-distance event-pairs that is adequate to meet our purpose.} to illuminate its knowledge transfer process. Through the section, we denote ``BERT'' and ``GNN'' as best-performing variants RoBERTa-Neighbor and RGAT, respectively.

\subsubsection{Model combination of MulCo}
Here we validate MulCo's utility as a single model to account for both short- and long-distance event pairs. We begin our comparison against a more straightforward way to form a joint model, e.g., ensembling BERT and GNN. We vary the cut-off we use to define short- and long-distance proximity bands in TDDAuto to evaluate how distance affects performance. This ensemble falls short of the theoretical “best case''~\cite{dong2020survey} by using all correct predictions from BERT and GNN that are respectively trained and evaluated on their assigned proximity band subset\footnote{For instance, BERT-Short denotes BERT baselines trained and evaluated only on the TDDAuto short-distance proximity band.}. Scores are shown in Table~\ref{perfect-ensemble-performance}. The observed under-performance suggests that, while event-pairs may encode temporal information from multiple proximity bands of the text, distance alone is not a reliable discriminator to define short- and long-distance pairs for joint models. MulCo, on the other hand, demonstrates a performance enhanced by collaborative knowledge integration between BERT and GNN, offering the potential for greater improvements by later joint models. 

\begin{table}[]
\caption{F1 score comparison between single models on short- and long-distance subset, perfect ensemble, and MulCo with distance cut-offs in TDDAuto. Note, the default distance cut-off is 2~\cite{naik2019tddiscourse}.}
\label{perfect-ensemble-performance}
\centering
\begin{tabular}{lllll}
\hline
\multicolumn{1}{l|}{\textbf{Distance-Cutoff}} & \multicolumn{1}{c|}{\textbf{2}} & \multicolumn{1}{c|}{\textbf{3}} & \multicolumn{1}{c|}{\textbf{4}} & \multicolumn{1}{c}{\textbf{5}} \\ \hline
\multicolumn{5}{c}{\textit{Performance on distance subset in TDDAuto}} \\ \hline
\multicolumn{1}{l|}{BERT-Short} & \multicolumn{1}{l|}{36.0} & \multicolumn{1}{l|}{60.8} & \multicolumn{1}{l|}{64.4} & 66.3 \\
\multicolumn{1}{l|}{GNN-Long} & \multicolumn{1}{l|}{43.8} & \multicolumn{1}{l|}{43.7} & \multicolumn{1}{l|}{44.1} & 43.8 \\ \hline
\multicolumn{1}{l|}{GNN-Short} & \multicolumn{1}{l|}{32.8} & \multicolumn{1}{l|}{42.1} & \multicolumn{1}{l|}{43.8} & 44.8 \\
\multicolumn{1}{l|}{BERT-Long} & \multicolumn{1}{l|}{70.7} & \multicolumn{1}{l|}{75.3} & \multicolumn{1}{l|}{73.5} & 62.0 \\ \hline
\multicolumn{5}{c}{\textit{Ensemble performance on TDDAuto}} \\ \hline
\multicolumn{1}{l|}{BERT-Short + GNN-Long} & \multicolumn{1}{c|}{42.3} & \multicolumn{1}{c|}{48.9} & \multicolumn{1}{c|}{52.4} & \multicolumn{1}{c}{55.3} \\
\multicolumn{1}{l|}{GNN-Short + BERT-Long} & \multicolumn{1}{c|}{63.4} & \multicolumn{1}{c|}{65.1} & \multicolumn{1}{c|}{59.7} & \multicolumn{1}{c}{51.8} \\ \hline
\multicolumn{1}{l|}{MulCo} & \multicolumn{1}{c|}{\textbf{77.1}} & \multicolumn{1}{c|}{\textbf{72.9}} & \multicolumn{1}{c|}{\textbf{72.9}} & \multicolumn{1}{c}{\textbf{66.8}} \\ \hline
\end{tabular}
\end{table}

\subsubsection{Knowledge aggregation of MulCo} 
\begin{figure}[htbp]
\centering
\centerline{\includegraphics[scale=0.06]{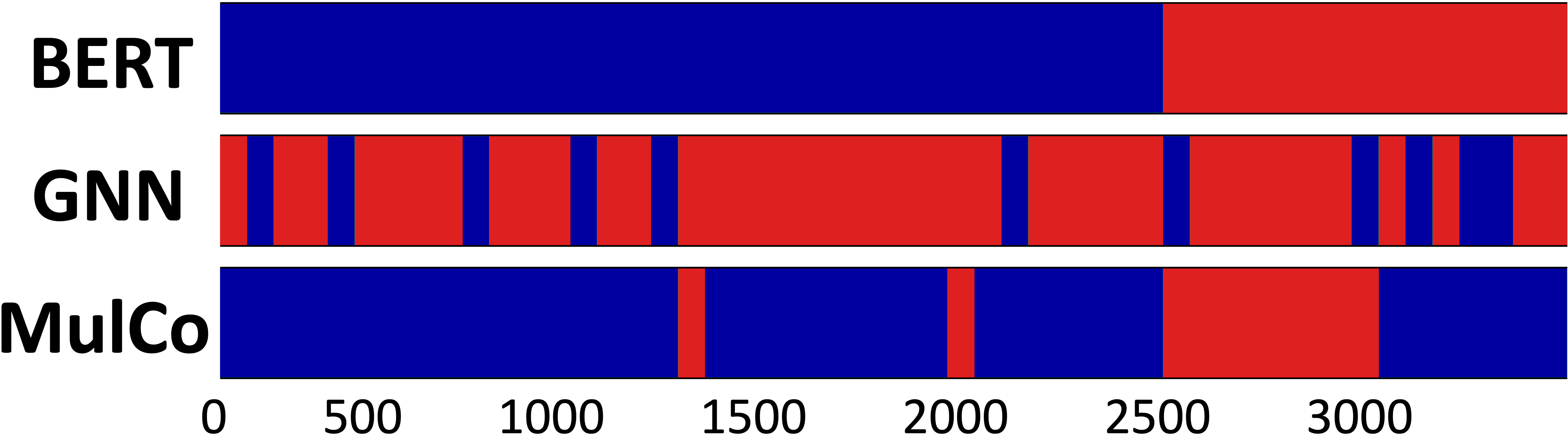}}
\caption{Correct (blue) and incorrect (red) predictions from BERT, GNN, and MulCo on TDDAuto. The x-axis denotes all test examples of TDDAuto indexed from 0 to 4258.} 
\label{distillation-preds}
\end{figure}

We argue that there are distinct proximity bands with different associated cues needed for full ETRE prediction, requiring an integrated approach. To validate that MulCo acts as an effective knowledge aggregator from BERT and GNN, we compare MulCo prediction with each on distinct event pairs. Figure~\ref{distillation-preds} shows that MulCo is able to perform predictions that BERT and GNN correctly perform, and correctly predicts several examples that are not captured by either BERT or GNN. This suggests MulCo is able to integrate and use knowledge effectively, better in some cases than either single model for overall more accurate prediction.

\subsubsection{Knowledge acquisition of MulCo} 
MulCo builds an integrated model that acquires temporal knowledge for short- and long-distance event pairs from BERT and GNN. To show the efficacy of MulCo knowledge distillation, we separately display the portion of predictions MulCo correctly makes which only one of BERT or GNN also makes (a ``unique prediction''). Unique predictions suggest a dependency on knowledge that is particular to one or the other proximity bands we study, and Table~\ref{distance-knowledge-distill} shows that MulCo retains that knowledge in over 50\% of all unique prediction types. Results further suggest that short-distance event pairs still benefit from GNN knowledge, as do long-distance with BERT. MulCo effectively incorporates the necessary insights those to achieve high performance across both short- and long-distance event pairs.

\begin{table}[]
\caption{Number and percentage of MulCo predictions that carry over from BERT and GNN unique prediction.}
\label{distance-knowledge-distill}
\centering
\begin{tabular}{lcc}
\hline
\multicolumn{1}{l|}{\textbf{Distance}} & \textbf{Short} & \textbf{Long} \\ \hline
\multicolumn{1}{l|}{TDDAuto Number of Test} & 877 & 2759 \\ \hline
\multicolumn{3}{c}{\textit{Unique Prediction}} \\ \hline
\multicolumn{1}{l|}{BERT} & 270 & 795 \\
\multicolumn{1}{l|}{GNN} & 22 & 83 \\ \hline
\multicolumn{3}{c}{\textit{MulCo Prediction in Unique Prediction}} \\ \hline
\multicolumn{1}{l|}{BERT} & 234 (87\%) & 710 (89\%) \\
\multicolumn{1}{l|}{GNN} & 12 (54\%) & 53 (64\%) \\ \hline
\end{tabular}
\end{table}

\subsection{Choice of backbone architecture}
\label{backbone-architecture}
\begin{table*}[]
\caption{Comparing F1 score with different BERT and GNN combination.}
\label{backbone-eval}
\centering
\begin{tabular}{l|ccc|ccc|ccc|ccc}
\hline
 & \multicolumn{3}{c|}{\textbf{TDDMan}} & \multicolumn{3}{c|}{\textbf{TDDAuto}} & \multicolumn{3}{c|}{\textbf{MATRES}} & \multicolumn{3}{c}{\textbf{TB-Dense}} \\ \hline
\textbf{Model} & \textbf{GCN } & \textbf{RGCN} & \textbf{RGAT} & \textbf{GCN } & \textbf{RGCN} & \textbf{RGAT} & \textbf{GCN } & \textbf{RGCN} & \textbf{RGAT} & \textbf{GCN } & \textbf{RGCN} & \textbf{RGAT} \\ \hline
BERT-Host & 40.9 & 41.0 & \textbf{47.3} & \textbf{70.3} & 66.3 & 66.2 & 84.9 & 84.2 & \textbf{85.7} & 82.0 & \textbf{83.5} & 81.4 \\
BERT-Neighbor & 41.1 & 38.4 & \textbf{48.1} & 63.1 & 64.7 & \textbf{68.3} & 82.7 & 84.2 & \textbf{86.0} & 79.0 & 78.2 & \textbf{81.0} \\
RoBERTa-Host & 46.4 & 47.7 & \textbf{51.5} & 67.2 & 75.2 & \textbf{76.0} & 86.0 & 88.6 & \textbf{89.7} & 79.2 & 82.8 & \textbf{84.2} \\
RoBERTa-Neighbor & 43.5 & 45.2 & \textbf{55.1*} & 74.8 & 66.5 & \textbf{77.1*} & 86.8 & 85.3 & \textbf{90.4*} & 80.5 & 84.7 & \textbf{85.6*} \\ \hline
\end{tabular}
\end{table*}

As discussed in Section~\ref{implementation}, different GNNs perform distinct actions, e.g., message passing and attention mechanisms. Some experimentation is necessary to identify the best target architecture for knowledge distillation. Table~\ref{backbone-eval} lists the F1 score of all combinations on different GNN and BERT baselines. From the evaluation, neighbor-sentence RoBERTa and RGAT achieve the best performance, which we attribute to their increased modeling capacity. The attention on relation type in RGAT has higher capacity to model complexity, providing extensive knowledge. Meanwhile, sentence dynamic masking~\cite{liu2019roberta} in RoBERTa provides more variance that can be regularized, allowing it to carry out robust knowledge distillation. Intuitively, injecting multiple sentences provides more contextual evidence for temporal relations, a cost-effective way to better extract temporal relations in both long- and short-distance scenarios.

\subsection{Performance analysis on k-hops and number of GNN layers}
\label{k-hops-performance}
A deeper GNN layer lets each node reach long-distance information. A larger k-hop neighborhood allows the node to capture complex local structures. Nevertheless, stacking more GNN layers can easily produce over-smoothing problem~\cite{oono2019graph}. Enlarging the neighborhood may introduce redundancy in local structure that is shared for neighborhood nodes. We examine different choices of k-hop and the number of layers on the best-performing architecture (RoBERTa-N + RGAT) and report F1 scores in Table~\ref{k-hops-layers}. We discover that more GNN layer and smaller neighborhood improves the performance for long-distance relation extraction, yet, it starts to drop when it goes more deeper. We surmise that hierarchical knowledge is more important than structural content on longer distance event pairs. Additionally, short-distance event pair doesn't require deeper and larger modeling space, as we find that increasing both layers and k-hop decreases the performance. Therefore, local structural knowledge is sufficient to characterize short-distance event pairs.

\begin{table}[htb]
\caption{F1 scores on different choices of k-hops and number of GNN layers. Note, we select the best performing architecture to perform the comparison.}
\label{k-hops-layers}
\centering
\begin{tabular}{lccccc}
\hline
\multicolumn{6}{c}{\textbf{TDDMan}} \\ \hline
\multicolumn{1}{l|}{\textbf{k-hops}} & \textbf{1-Layer} & \textbf{2-Layer} & \textbf{3-Layer} & \textbf{4-Layer} & \textbf{5-Layer} \\ \hline
\multicolumn{1}{l|}{1} & 47.7 & \textbf{55.1} & 49.8 & 49.2 & 48.4 \\
\multicolumn{1}{l|}{2} & 49.1 & 52.3 & 50.1 & 44.5 & 48.6 \\
\multicolumn{1}{l|}{3} & 47.5 & 48.6 & 47.5 & 45.4 & 48.7 \\
\multicolumn{1}{l|}{4} & 49.6 & 50.8 & 47.1 & 44.0 & 47.9 \\
\multicolumn{1}{l|}{5} & 50.7 & 50.9 & 48.8 & 46.4 & 45.8 \\ \hline
\multicolumn{6}{c}{\textbf{TDDAuto}} \\ \hline
\multicolumn{1}{l|}{\textbf{k-hops}} & \textbf{1-Layer} & \textbf{2-Layer} & \textbf{3-Layer} & \textbf{4-Layer} & \textbf{5-Layer} \\ \hline
\multicolumn{1}{l|}{1} & 67.4 & 71.6 & 66.6 & 69.7 & 68.3 \\
\multicolumn{1}{l|}{2} & 67.9 & 69.9 & \textbf{77.1} & 70.7 & 69.8 \\
\multicolumn{1}{l|}{3} & 68.0 & 71.7 & 69.3 & 68.4 & 65.1 \\
\multicolumn{1}{l|}{4} & 68.1 & 67.2 & 67.4 & 64.9 & 72.3 \\
\multicolumn{1}{l|}{5} & 68.1 & 73.4 & 74.6 & 68.2 & 72.1 \\ \hline
\multicolumn{6}{c}{\textbf{MATRES}} \\ \hline
\multicolumn{1}{l|}{\textbf{k-hops}} & \textbf{1-Layer} & \textbf{2-Layer} & \textbf{3-Layer} & \textbf{4-Layer} & \textbf{5-Layer} \\ \hline
\multicolumn{1}{l|}{1} & \textbf{90.4} & 87.9 & 89.7 & 86.8 & 85.7 \\
\multicolumn{1}{l|}{2} & 87.9 & 85.7 & 89.7 & 85.3 & 87.1 \\
\multicolumn{1}{l|}{3} & 87.5 & 86.0 & 88.2 & 85.7 & 86.8 \\
\multicolumn{1}{l|}{4} & 88.2 & 85.3 & 87.5 & 87.9 & 86.0 \\
\multicolumn{1}{l|}{5} & 86.4 & 86.8 & 87.5 & 86.8 & 86.0 \\ \hline
\multicolumn{6}{c}{\textbf{TB-Dense}} \\ \hline
\multicolumn{1}{l|}{\textbf{k-hops}} & \textbf{1-Layer} & \textbf{2-Layer} & \textbf{3-Layer} & \textbf{4-Layer} & \textbf{5-Layer} \\ \hline
\multicolumn{1}{l|}{1} & \textbf{85.5} & 80.6 & 83.1 & 82.3 & 83.1 \\
\multicolumn{1}{l|}{2} & 83.4 & 82.8 & 80.4 & 79.5 & 83.5 \\
\multicolumn{1}{l|}{3} & 81.9 & 82.8 & 76.8 & 81.7 & 80.1 \\
\multicolumn{1}{l|}{4} & 83.0 & 82.9 & 81.4 & 81.1 & 80.4 \\
\multicolumn{1}{l|}{5} & 83.3 & 80.7 & 79.0 & 78.8 & 84.0 \\ \hline
\end{tabular}
\end{table}

\subsection{Discussion of performance on Graph Neural Network baselines}
\label{gnn-performance}
To further discuss the sub-optimal performance of Graph Neural Networks (GNNs), we select TIMERS~\cite{mathur2021timers} and UCGraph~\cite{liu2021discourse}, the current SOTA GNN-based methods to evaluate on long- and short-distance dataset, e.g., TDDMan, TDDAuto, and Timebank-Dense, to analyze their GNN components. We directly cite the evaluation scores from the experimental section in their original paper. We compare three models: (\textbf{1}) UCGraph-RGCN; a powerful RGCN that utilizes novel uncertainty modeling (refer to original paper~\cite{liu2021discourse} for more details), (\textbf{2}) GR-GCN; a vanilla RGCN with gating mechanism to control feature propagation through layers (more details in~\cite{mathur2021timers}), and (\textbf{3}) TIMRES; combining GR-GCN and BERT context encoder to form an integrated model for event temporal ordering (refer~\cite{mathur2021timers} for more details). As we can see in Table~\ref{gnn-base}, UCGraph can obtain high performance on TDDMan and be competitive on TDDAuto. Since TDDMan contains only long-distance event pairs, it shows that GNN-based method has the potential to model long-distance event pairs. However, when performing on a short-distance dataset, e.g., TB-Dense, it performs weakly compared to pre-trained language model baselines (BERT and RoBERTa). It is obvious that BERT-based models perform well on short-distance event pairs, which powerful GNN usually does not. On the other hand, it is evident that TIMERS performance mainly depends on its BERT context encoding. The vanilla RGCN is incapable to capture both long- and short-distance context, for its poor evaluation results. When combining them, the performance substantially improves. It tells us two things: \textbf{(1)} either GNN needs to be more powerful by including other modeling concepts (e.g., uncertainty), \textbf{(2)} or combine with BERT-based language model to implicitly provide long-distance context information. Obviously, GNN does improve long-distance modeling but performs sub-optimal on all short-distance cases.

\begin{table}[]
\caption{F1 score comparison under different GNN baselines. We directly cite evaluation scores from their original paper~\cite{mathur2021timers, liu2021discourse}.}
\label{gnn-base}
\centering
\begin{tabular}{l|cccc}
\hline
\textbf{Model} & \textbf{TDDMan} & \textbf{TDDAuto} & \textbf{MATRES} & \textbf{TB-Dense} \\ \hline
BERT & 37.5 & 62.3 & 78.1 & 62.2 \\
RoBERTa & 37.1 & 61.6 & 78.4 & 61.9 \\
UCGraph & 43.4 & 61.2 & - & 59.1 \\
GR-GCN & 33.7 & 51.6 & 68.6 & 50.6 \\
TIMERS & 45.5 & 71.1 & 82.3 & 67.8 \\ \hline
\end{tabular}
\end{table}

\begin{table}[]
\caption{Comparing F1 score with allowing \textbf{A} or stopping \textbf{S} gradient through contrastive objective.}
\label{allow-stop-eval}
\centering
\begin{tabular}{l|cc|cc|cc|cc}
\hline
 & \multicolumn{2}{c|}{\textbf{B-H}} & \multicolumn{2}{c|}{\textbf{B-N}} & \multicolumn{2}{c|}{\textbf{R-H}} & \multicolumn{2}{c}{\textbf{R-N}} \\ \hline
\textbf{Dataset} & \textbf{A} & \textbf{S} & \textbf{A} & \textbf{S} & \textbf{A} & \textbf{S} & \textbf{A} & \textbf{S} \\ \hline
TDDMan & 36.2 & \textbf{47.3} & 38.8 & \textbf{49.1} & 42.2 & \textbf{51.5} & 51.3 & \textbf{55.1} \\
TDDAuto & 57.5 & \textbf{70.3} & 65.5 & \textbf{68.3} & 67.5 & \textbf{76.0} & 68.1 & \textbf{77.1} \\
MATRES & 83.1 & \textbf{85.7} & 84.9 & \textbf{86.0} & 87.1 & \textbf{89.7} & 86.2 & \textbf{90.4} \\
TB-Dense & 79.5 & \textbf{83.5} & 75.9 & \textbf{81.0} & 83.4 & \textbf{84.2} & 78.1 & \textbf{85.6} \\ \hline
\end{tabular}
\end{table}

\subsection{Performance impact of stop-gradient and interpretation} 
\label{stop-gradient}
Stop-gradient~\cite{chen2021exploring} is a function that removes gradient flows to pass through its input. We evaluate the F1 scores if allowing gradient flows passing through the second input of $\ell_\text{cl}$ in CoD loss (Eq.~\ref{cod}):
\begin{equation}
    \mathcal{L}_{\text{CoD}_{\textbf{A}}} = \sum^N_{i=1}\ell_\text{cl}(\bm{h^\text{bert}_i},\bm{\mathcal{H}_i}) + \ell_\text{cl}(\bm{\mathcal{H}_i},\bm{h^\text{bert}_i})
\end{equation}
\noindent
where we remove the $\hat{\cdot}$ from $\bm{\mathcal{H}_i}$ and $\bm{h^\text{bert}_i}$, the second term in $\ell_\text{cl}$. Table~\ref{allow-stop-eval} shows that ``Passing'' through gradient flows on all BERT baselines ``defeats'' the performance on all datasets.  This finding validates the necessity of using a stop-gradient. To simplify our interpretation, let $F_{\theta}, G_{\pi}$ denote BERT and GNN representations parameterized by trainable weights $\theta$ and $\pi$. Let $\hat{\cdot}$ be a stop gradient operator. Following the derivation from~\cite{chen2021exploring}, we now cast the contrastive learning task Eq.\ref{loss-cl} into two subtasks:
\begin{align}
    \min_{\theta, \pi} \ell_\text{cl}(F_{\theta}, \hat G_{\pi}) & \approx \min_{\theta} \mathrm{sim}(F_{\theta}, G) \label{stop-g} \\
    \min_{\pi, \theta} \ell_\text{cl}(G_{\pi}, \hat F_{\theta}) & \approx \min_{\pi} \mathrm{sim}(G_{\pi}, F) \label{stop-f}
\end{align}
\noindent
where $\mathrm{sim}(\cdot)$ denotes similarity, and $G,F$ denote variables that gradients are removed from $\pi,\theta$ that associate with the original $G_{\pi},F_{\theta}$. Mutual distillation, DML~\cite{zhang2018deep}, requires solving the right terms in Eq.~\ref{stop-g} first, and then~\ref{stop-f}. When computing the joint loss (summing up both left terms in Eq.~\ref{stop-g} and~\ref{stop-f}), we are actually solving mutual distillation objectives, as mentioned earlier, which maximizes the similarity for $F_{\theta}$ given the embedding from $G$ as the target objective and vice versa. To this end, the alternative optimization is now formulated as a single end-to-end training. \\

\section{Conclusion}
Learning to leverage temporal cues on long- and short-distance event pairs is a crucial step towards robust temporal relation extraction, yet progress has been hampered by limitations in transformer context lengths. With our simple, effective, and high-performing MulCo approach, multi-scale knowledge co-distillation overcomes BERT's long-distance barriers, and simultaneously optimizes knowledge, representation, and distillation, thus leading to new state-of-the-art performance on the event temporal relation extraction task.

\begin{acks}
This research was supported, in part, by the Intramural Research Program of the National Institutes of Health and the US Social Security Administration.
\end{acks}

\appendix
\section{Appendix}
\label{sec:Appendix}
\subsection{Dataset IAA vs Performance}
\label{appendix-iaa}
To argue the potential question regarding whether it's possible to have high-performance model that exceeds the Inter-Annotator Agreement (IAA) for benchmark datasets, it should be noted that the IAA for these datasets does not constitute an upper bound on model performance. The temporal order of two real events is an objective property, even if human annotators find it difficult to deduce and thus disagree heavily. Additionally, all benchmarks (TDDiscourse~\cite{naik2019tddiscourse}, TimeBank-Dense~\cite{cassidy2014annotation}, and MATRES~\cite{ning2018multi}) have undergone post-IAA adjudication, and these adjudicated labels can help models identify features that improve performance beyond IAA.

\subsection{F1 score details on TimeBank-Dense and MATRES}
\label{appendix-f1}
To be consistent with past work on this dataset, The precision calculation for TimeBank-Dense (TB-Dense) differs from  standard precision. At test time, models label relations for event pairs sampled per some heuristic (e.g., all event pairs occurring within 2 sentences of each other) since labeling all possible event pairs gets expensive for longer documents. Sometimes the set of pairs labeled by a model is a strict superset of the gold data. In this situation, model predictions for pairs that are not labeled in the gold standard are not included while computing precision (see lines 336-339 in implementation from TB-Dense authors~\cite{Nchambers}.), and Precision / Recall/ F1 scores all end up being the same, as in prior work (refer Table 7 in [2]). Additionally, marginal improvements on TDDAuto vs other datasets is not too surprising because we already expected improving performance on mixed datasets to be a greater challenge, given most recent SOTA work has not been able to solve this issue well. We have also submitted our source code to support our experimental scores. 


\bibliographystyle{ACM-Reference-Format}
\bibliography{reference}

\end{document}